%% file: neurips_2026.tex
\documentclass{article}

\PassOptionsToPackage{numbers, compress}{natbib}
\usepackage[preprint]{neurips_2026}

\usepackage[utf8]{inputenc} 
\usepackage[T1]{fontenc}    
\usepackage{hyperref}       
\usepackage{url}            
\usepackage{booktabs}       
\usepackage{amsfonts}       
\usepackage{nicefrac}       
\usepackage{microtype}      
\usepackage[table]{xcolor}  
\usepackage{algorithm}
\usepackage{algorithmic}
\usepackage{graphicx}
\usepackage{caption}
\usepackage{float}
\usepackage{amsmath}
\usepackage{multirow}
\usepackage{graphicx}
\usepackage{makecell}
\usepackage{wrapfig}

\usepackage{listings}

\title{CHASD: Language Increment-Calibrated Contrastive Decoding against Hallucination in LVLMs}

\newcommand{\up}{\textcolor{green!60!black}{$\uparrow$}}
\newcommand{\down}{\textcolor{red!70!black}{$\downarrow$}}

\author{
  Xiaoyi Huang$^{1}$ \quad
  Kejia Zhang$^{2}$ \quad
  Zhiming Luo$^{2}$ \\
  $^{1}$Institute of Artificial Intelligence, Xiamen University \\
  $^{2}$Department of Artificial Intelligence, Xiamen University
}

\begin{document}

\maketitle

\begin{abstract}
Large Vision-Language Models have shown strong multimodal reasoning capabilities, yet they remain susceptible to object hallucinations when language priors dominate insufficient or misaligned visual evidence. Training-free contrastive decoding methods mitigate this issue by comparing predictions from original and perturbed visual inputs, but existing approaches either apply global perturbations that may alter useful visual evidence or invoke an additional negative branch at every decoding step. In this paper, we observe that hallucination risks are \textbf{transient} and \textbf{token-specific}: visual attention shifts across generated tokens, while some functional tokens are produced with high confidence and do not require contrastive calibration. Based on this observation, we propose \textbf{Contrastive Hallucination-Aware Step-wise Decoding (CHASD)} for Large Vision-Language Models, an inference-time framework for \textbf{"calibration on demand"}. CHASD uses an uncertainty-driven confidence gate to activate the contrastive branch only when the maximum probability of the next-token is less than the threshold, and constructs the negative branch through attention-guided localized perturbations of the currently salient visual tokens. This design reduces unnecessary negative-branch forward passes while preserving the original distribution for high-confidence steps. Experiments on POPE, AMBER, MME, MMHal-Bench, and CHAIR show that CHASD improves hallucination-related metrics over strong training-free baselines with competitive inference efficiency.
\end{abstract}

\input{Sections/1.introduction}
\input{Sections/2.relatedwork}
\input{Sections/3.methodology}

\input{Sections/4.experiments}
\input{Sections/5.conclusion}

\bibliography{reference}
\bibliographystyle{abbrv}



\appendix
\input{Sections/appendix}

\end{document}

%% file: Sections/1.introduction.tex
\section{Introduction}

Large Vision-Language Models (LVLMs) have achieved remarkable success in bridging visual perception with linguistic reasoning~\cite{shikra,qwen,minigpt}. However, they remain susceptible to object hallucinations, wherein the models generate descriptions of non-existent objects or incorrect relationships~\cite{guan,tong,zhou}. Recent studies indicate that this issue primarily stems from an over-reliance on language priors~\cite{chen,tong, wu}; when visual evidence is ambiguous or attention misaligned, LVLMs tend to produce high-frequency token sequences that override the actual visual context.

To mitigate this without retraining costs, training-free contrastive decoding (CD)~\cite{ED,Li,ICD} has emerged as a promising paradigm. The pioneering VCD~\cite{vcd} amplifies visual signals by contrasting predictions from original and globally blurred images. Despite its effectiveness, the undifferentiated global perturbation in VCD can also modify useful visual features. Subsequent efforts have sought to refine this via more granular interventions. Specifically, AvisC~\cite{avisc} identifies irrelevant visual tokens through hierarchical attention analysis to enable localized processing. Nevertheless, its negative sample is \textbf{static}, failing to adapt to the dynamic shift of visual focus during the decoding process. Conversely, works such as SID~\cite{sid} and HICD~\cite{hicd} recognize the \textbf{transient} nature of hallucinations, employing context-aware or incremental strategies to generate token-specific negative examples. However, these methods essentially require \textbf{dual forward passes} for all tokens in a sequence, which increases the computational burden and inference latency.

\begin{figure}[t]
\centering
\includegraphics[   width=\linewidth,
    trim=0 4cm 0 2cm,
    clip]{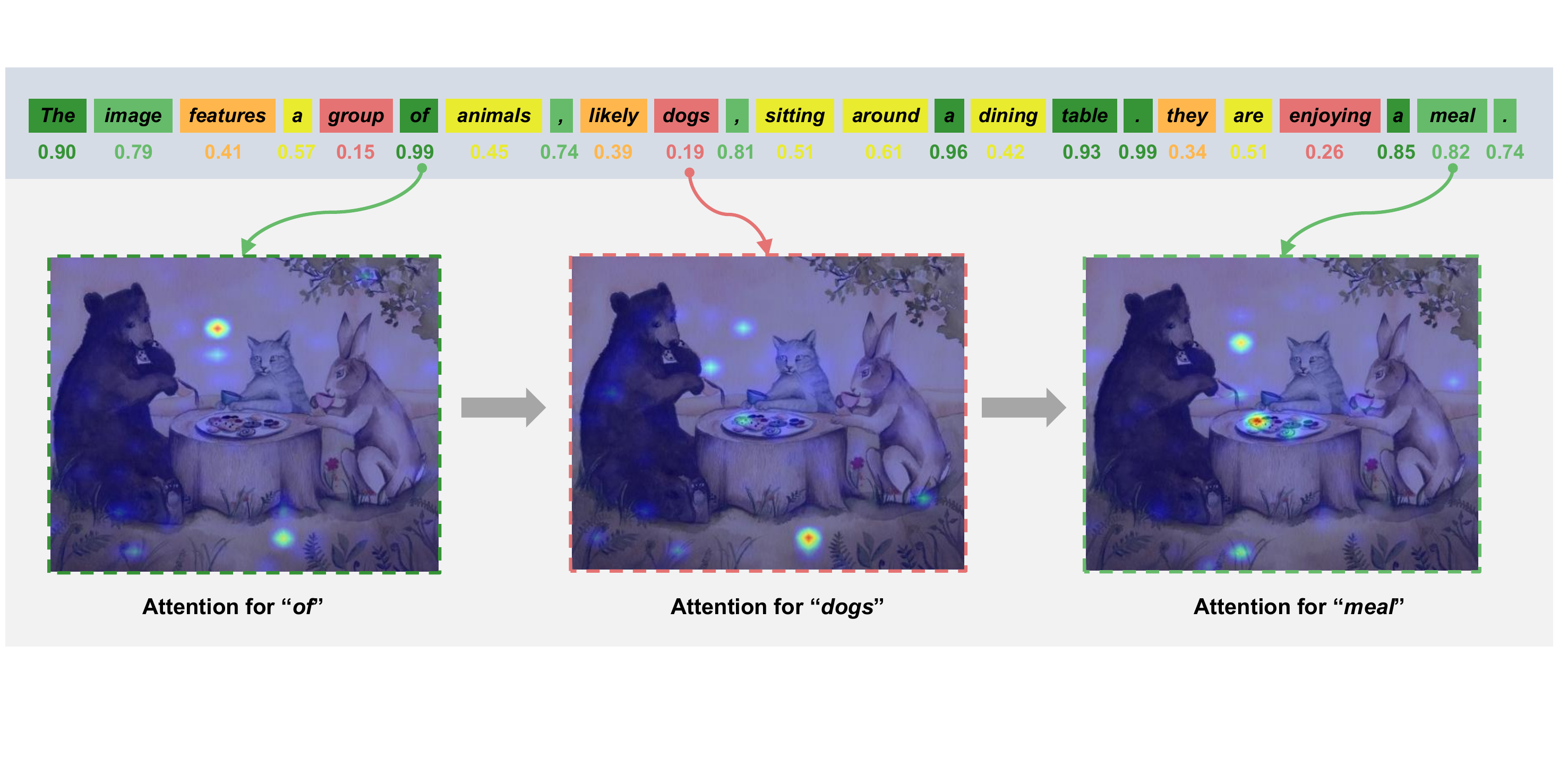}
\caption{Visualization of \textbf{step-wise probabilities} and \textbf{cross-attention maps} during the LVLM generation process. \textbf{Top:} The model assigns varying confidence scores to each token. The high confidence observed in functional tokens suggests that the contrastive decoding branch can be \textbf{selectively bypassed} to enhance inference efficiency. \textbf{Bottom:} The model’s visual focus areas shift dynamically when generating different tokens. This variation highlights the necessity of constructing negative samples on a \textbf{token-by-token} basis to ensure precise spatial alignment.}
\label{fig:motivation}
\end{figure}

Our motivation stems from the empirical observation that hallucination risks are highly transient and token-specific. As illustrated in Figure.~\ref{fig:motivation}, the visual regions attended by the model vary across tokens, necessitating a dynamic formulation of negative samples. Furthermore, prediction confidence exhibits extreme variance: functional or syntactic tokens (e.g., "\textit{of}") often present near-unity confidence and diffused attention, indicating they are driven by stable linguistic logic. In contrast, hallucination-prone entities (e.g., "\textit{dogs}") exhibit lower confidence coupled with spurious visual anchors. This suggests that applying a continuous, sequence-level contrastive penalty is both computationally redundant and theoretically sub-optimal.

Driven by these insights, we propose that effective hallucination suppression requires \textbf{"calibration on demand"}. We introduce
\textbf{C}ontrastive \textbf{H}allucination-\textbf{A}ware \textbf{S}tep-wise \textbf{D}ecoding (\textbf{CHASD}) for Large Vision-Language Models, an inference framework governed by a \textbf{spatio-temporal joint constraint}. Temporally, CHASD utilizes an uncertainty-driven confidence gating mechanism to determine \textit{when} to intervene, activating the contrastive branch only when the maximum probability of the next-token falls below the threshold. Spatially, CHASD identifies the most salient visual tokens based on cross-modal attention weights to determine \textit{where} to perturb the visual evidence for the negative branch. By applying localized perturbations to these specific visual anchors only during vulnerable decoding steps, CHASD reduces unnecessary negative-branch forward passes while preserving the original distribution for high-confidence text.

The primary contributions of this work are summarized as follows:
\begin{itemize}
    \item We demonstrate that hallucination risks are token-specific rather than uniformly distributed across the sequence. We reveal that contrastive decoding is only essential for specific "vulnerable" steps, providing a empirical basis for efficiency optimization.
    \item We propose \textbf{CHASD}, a training-free framework that introduces a spatio-temporal joint constraint. It dynamically couples \textbf{temporal confidence gating} with \textbf{spatial attention guidance} to calibrate only selected low-confidence decoding steps.
    \item Extensive evaluations across POPE~\cite{pope}, AMBER~\cite{amber}, MME~\cite{mme}, MMHal-Bench~\cite{mmh} and CHAIR~\cite{chair} show that CHASD improves hallucination-related metrics over strong training-free baselines. It reduces unnecessary negative-branch forward passes while maintaining competitive generation quality.
\end{itemize}

%% file: Sections/2.relatedwork.tex
\section{Related Work}

\subsection{Object Hallucination and Mitigation Strategies}
Object hallucination~\cite{guan, zhang2,zhang} remains a primary impediment to the reliable deployment of LVLMs. This phenomenon typically originates from data-level biases, an over-reliance on language priors~\cite{chen,tong, wu}, and the model's inherent difficulty in grounding real-world contexts~\cite{Ji}. Early mitigation efforts primarily focused on \textbf{training-based approaches}~\cite{feng, mmh, zhou} and \textbf{data optimization}~\cite{ma,you,  yue}. However, these methods often incur prohibitive computational overhead and time costs, while risking the degradation of the model's general-purpose capabilities.

To bypass the cost of retraining, \textbf{training-free methods} have gained significant attention. For instance, OPERA~\cite{opera} alleviates hallucinations through an over-trust penalty and a rollback allocation strategy. Recently, \textbf{contrastive decoding}~\cite{Li} has emerged as a dominant paradigm. The pioneering work, VCD~\cite{vcd}, amplifies visual signals by contrasting predictions from original and blurred images. Building upon this, the research community has proposed various strategies for negative sample construction, such as using sub-image cropping~\cite{ED}, blank images~\cite{m3id}, hallucination visualization~\cite{convis} and visual token shuffling~\cite{SDCD}. Other advancements focus on optimizing the CD process itself, including fuzzy reasoning~\cite{fuzzyCD}, decider-based selection strategies~\cite{ vacode,octopus} and instruction-based contrastive decoding~\cite{ICD}.

\subsection{Fine-grained Contrastive Decoding}
Despite its milestone status in hallucination mitigation, VCD suffers from performance and efficiency bottlenecks due to its fixed global perturbation strategy and full-sequence contrastive decoding. Subsequent research has sought to improve upon these coarse-grained characteristics by leveraging internal model states to implement more targeted interventions.

In the spatial dimension, AvisC~\cite{avisc} identifies blind tokens through hierarchical attention analysis to enable localized processing. However, its negative sample construction remains inherently \textbf{static}, making it difficult to adapt to the dynamic shift of visual focus during the generation process. Conversely, some studies recognize the \textbf{transient nature} of hallucinations and employ various methods to alleviate them: SID~\cite{sid} develops a context-aware and text-aware token selection strategy to amplify hallucinations via progressively chosen irrelevant tokens, while HICD~\cite{hicd} constructs negative samples by step-wise selecting and shuffling key attention heads. Nonetheless, these methods essentially tend to perform dual calculations for every token in the sequence, which severely increases the computational burden and inference latency.

In summary, existing fine-grained methods treat spatial localization and temporal selection as isolated problems, leading to an imbalance between spatial precision and temporal efficiency. Our proposed \textbf{CHASD} addresses this limitation by introducing a \textbf{spatio-temporal coupled constraint}. By ensuring that contrastive decoding is performed only \textit{when} necessary (temporal dimension) and only \textit{where} relevant (spatial dimension), CHASD reduces unnecessary negative-branch computation while retaining localized visual calibration.

%% file: Sections/3.methodology.tex
\section{Methodology}

\begin{figure}[t]
\centering
\captionsetup{type=figure}
\vspace{2pt}
\includegraphics[width=1\linewidth]{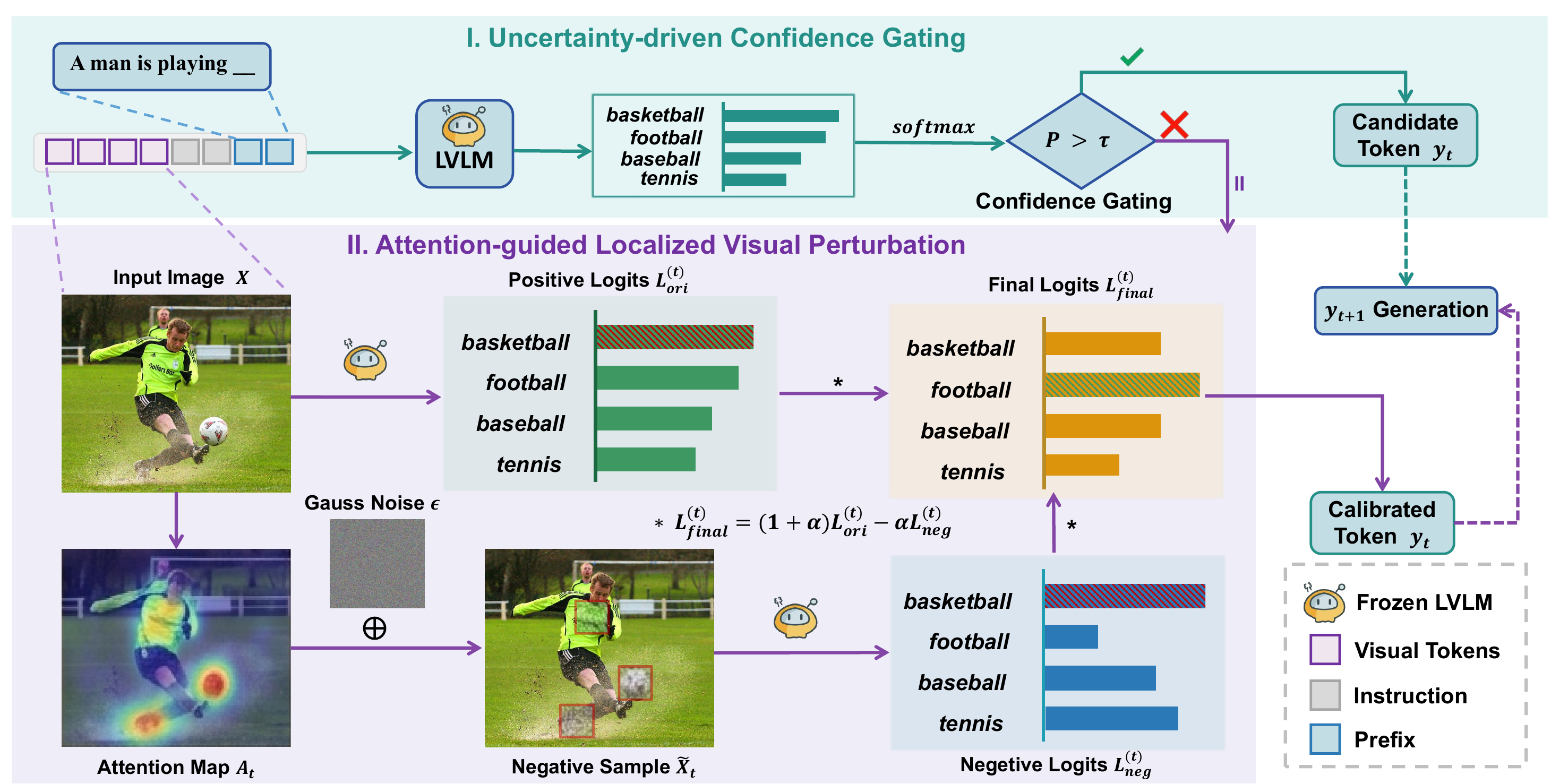}
\caption{\textbf{Overview of the CHASD at time step $t$.} For the initial token generated at the current step, it first undergoes 
    \textbf{(I) Uncertainty-driven Confidence Gating}: if the maximum predictive probability $P$ exceeds the threshold $\tau$, the system directly outputs the candidate token.
    Otherwise, the system enters the \textbf{(II) Attention-guided Localized Visual Perturbation} branch, leveraging an attention map to identify salient regions and constructing a negative sample via localized Gaussian noise injection. By contrasting the resulting negative logits with the original distribution, the system produces the final calibrated token.}
\label{fig:pipeline}
\end{figure}

\subsection{Preliminaries}

\textbf{Token Generation in LVLMs.} 
LVLMs typically employ an autoregressive decoding strategy to bridge visual and linguistic modalities. Formally, given an input image $X$ and a prefix sequence $Y_{<t}$ (comprising task instructions and previously generated tokens), the model $f_\theta$ computes the probability of the next token $y_t$ by applying a Softmax normalization over the original logits:
\begin{equation}
P(y_t \mid Y_{<t}, X) = \text{Softmax}(\mathbf{L}_{ori}^{(t)})[y_t], \quad \text{where} \quad \mathbf{L}_{ori}^{(t)} = f_\theta(Y_{<t}, X)
\end{equation}
During decoding, the model iteratively samples tokens to construct coherent responses. However, this process is highly susceptible to language priors. The model often relies excessively on statistical patterns learned during pre-training rather than the factual visual evidence provided in $X$, intrinsically leading to object hallucinations.

\textbf{VCD Framework.} 
To mitigate this bias, Visual Contrastive Decoding (VCD) \cite{vcd} introduces a calibration mechanism that penalizes hallucination-prone distributions. Specifically, VCD calculates a negative logit distribution $\mathbf{L}_{neg}^{(t)}$ based on a statically corrupted image $X'$ (usually by applying global noise), and formulates the calibrated logits as:
\begin{equation}
\mathbf{L}_{CD}^{(t)} = (1 + \alpha) \cdot \mathbf{L}_{ori}^{(t)} - \alpha \cdot \mathbf{L}_{neg}^{(t)}
\end{equation}
where $\alpha$ denotes the contrastive strength. Furthermore, to prevent the contrastive operation from amplifying irrelevant tokens, VCD utilizes the Adaptive Probabilistic Constraint (APC), which truncates the sampling space to a candidate set $\mathcal{V}_t$:
\begin{equation}
\mathcal{V}_t = \{ w \in \mathcal{W} \mid P(w \mid Y_{<t}, X) > \beta \cdot \max_{w' \in \mathcal{W}} P(w' \mid Y_{<t}, X) \}
\end{equation}
While APC effectively circumvents semantic collapse, the construction of negative samples in VCD relies entirely on static, global noise injection. This limits its capacity to perceive and respond to the dynamically drifting visual attention of the model during decoding, thereby motivating our exploration of a fine-grained, dynamic calibration mechanism.

\input{Tables/algorithm}
\subsection{Contrastive Hallucination-Aware Step-wise Decoding}
To address the limitations of existing contrastive decoding methods in capturing transient attentional shifts, we propose the \textbf{C}ontrastive \textbf{H}allucination-\textbf{A}ware \textbf{S}tep-wise \textbf{D}ecoding (\textbf{CHASD}) for Large Vision-Language Models  framework. CHASD transitions the hallucination calibration strategy from static, global perturbations to a dynamic, step-wise localized intervention. The execution of CHASD is comprised of two core modules:

\subsubsection{Uncertainty-driven Confidence Gating}
To mitigate computational redundancy and prevent over-calibration of high-confidence predictions—such as common functional words or unambiguous visual entities—we introduce a confidence gating mechanism (Stage I in Figure~\ref{fig:pipeline}). At decoding step $t$, the gate is applied to the original next-token distribution before sampling. The contrastive calibration process is activated only if the maximum prediction probability of this distribution falls below a predefined threshold $\tau$:
\begin{equation}
\mathbf{L}_{final}^{(t)} = 
\begin{cases} 
\mathbf{L}_{CD}^{(t)}, & \text{if } \max_{v \in \mathcal{V}} P_{ori}(v | Y_{<t}, X) < \tau \\
\mathbf{L}_{ori}^{(t)}, & \text{otherwise}
\end{cases}
\end{equation}
where $\mathcal{V}$ denotes the vocabulary. This mechanism ensures that CHASD concentrates the additional negative-branch computation on steps exhibiting high uncertainty, thereby suppressing hallucinations while preserving the model's intrinsic generation capability at high-confidence steps.

\subsubsection{Attention-guided Localized Visual Perturbation}
If the next-token distribution fails the confidence gate, the spatial calibration process is triggered (Stage II in Figure~\ref{fig:pipeline}). At each triggered decoding step $t$, we extract the cross-modal attention weights from the model's last layer, denoted as $\mathbf{A}_t \in \mathbb{R}^{H \times N}$. By averaging across the $H$ attention heads, we calculate the saliency score $s_i^{(t)}$ for each visual token:
\begin{equation}
s_i^{(t)} = \frac{1}{H} \sum_{h=1}^{H} A_{h,i}^{(t)}, \quad i \in \{1, \dots, N\}
\end{equation}
The vector $\mathbf{s}^{(t)}$ characterizes the model's instantaneous reliance on specific visual regions during the current token prediction. Unlike global blinding techniques, CHASD implements localized perturbation based on these saliency scores. We define $k \in (0,1]$ as the attention sparsity ratio and select $m=\lceil kN \rceil$ visual tokens at each triggered step. The dynamic binary mask $\mathcal{M}_t$ anchors these $m$ most salient visual tokens:
\begin{equation}
\mathcal{M}_t(i) = \mathbb{I}(s_i^{(t)} \in \text{Top-}m(\mathbf{s}^{(t)}, m)), \quad m=\lceil kN \rceil
\end{equation}
Subsequently, Gaussian noise is injected exclusively into these high-attention regions to dynamically construct the negative sample $\tilde{X}_t$ for step $t$:

\begin{equation}
\tilde{X}_{t} = X \odot (1 - \mathcal{M}_{t}') + (X + \epsilon) \odot \mathcal{M}_{t}'
\end{equation}
where $\sigma$ is a fixed standard Gaussian perturbation scale and $\mathcal{M}_t'$ is upsampled from the 1D token mask $\mathcal{M}_t$. The resulting negative logits $\mathbf{L}_{neg}^{(t)} = f_\theta(Y_{<t}, \tilde{X}_t)$ allow the model to compare its original next-token distribution with a distribution obtained after perturbing the currently attended visual evidence. Detailed algorithms are provided in Algorithm~\ref{alg:dvcd}.

%% file: Tables/algorithm.tex
\begin{algorithm}[t]
\caption{The Inference Procedure of CHASD}
\label{alg:dvcd}
\begin{algorithmic}[1]
\REQUIRE Image $X$, Prefix sequence $Y_{<t}$, Contrastive strength $\alpha$, Gating threshold $\tau$, Attention sparsity ratio $k$, Gaussian perturbation scale $\sigma$.
\ENSURE Generated sequence $Y$
\STATE $Y \leftarrow \emptyset$, $t \leftarrow 1$
\WHILE{$y_t \neq \text{[EOS]}$}

    \STATE Obtain original logits: $\mathbf{L}_{ori}^{(t)} \leftarrow f_\theta(Y_{<t}, X)$ \hfill\textcolor{gray}{//Original Logits Generation}
    \STATE $C_t \leftarrow \max(\text{Softmax}(\mathbf{L}_{ori}^{(t)}))[y_t]$ 
    
    \IF{$C_t < \tau$}
        
        \STATE $\mathbf{A}_t \leftarrow \text{Extract last-layer cross-attention weights}$\hfill\textcolor{gray}{//Attention-Guided Saliency Modeling}
        \STATE $s_i^{(t)} = \frac{1}{H} \sum_{h=1}^{H} A_{h,i}^{(t)}, \quad i \in \{1, \dots, N\}$

        \STATE $m \leftarrow \lceil kN \rceil$
        \STATE $\mathcal{M}_t' \leftarrow \text{SpatialMap}( \mathbb{I}(s_i^{(t)} \in \text{Top-}m(\mathbf{s}^{(t)}, m)))$ \hfill\textcolor{gray}{//Localized Dynamic Perturbation}
     
        \STATE $\tilde{X}_t \leftarrow X \odot (\mathbf{1} - \mathcal{M}_{t}') + (X + \epsilon) \odot \mathcal{M}_{t}', \quad \epsilon \sim \mathcal{N}(0, \sigma^2 \mathbf{I})$
        \STATE $\mathbf{L}_{neg}^{(t)} \leftarrow f_\theta(Y_{<t}, \tilde{X}_t)$
        \STATE $\mathbf{L}^{(t)} \leftarrow (1+\alpha)\mathbf{L}_{ori}^{(t)} - \alpha\mathbf{L}_{neg}^{(t)}$
    \ELSE
        \STATE $\mathbf{L}^{(t)} \leftarrow \mathbf{L}_{ori}^{(t)}$
    \ENDIF
    
    \STATE $\mathcal{V}_t = \{ w \in \mathcal{W} \mid P(w \mid Y_{<t}, X) > \beta \cdot \max_{w' \in \mathcal{W}} P(w' \mid Y_{<t}, X) \}$
    \STATE $y_t \leftarrow \text{Sample from } \mathbf{L}^{(t)} \text{ restricted to } \mathcal{V}_t$
    \STATE $Y \leftarrow Y \cup \{y_t\}$ 
    \STATE $t \leftarrow t + 1$
\ENDWHILE
\RETURN $Y$
\end{algorithmic}
\end{algorithm}

%% file: Sections/4.experiments.tex
\section{Experiments}

\subsection{Experimental Setups}

\textbf{Models and Hyperparameters.} Following the experimental protocols of existing mainstream literature, we adopt two representative LVLMs as our evaluation backbones: LLaVA-1.5-7B~\cite{llava} and InstructBLIP-7B~\cite{instructblip} (scalability analyses on more different models are provided in Appendix~\ref{sec:more_results}). For the hyperparameters of CHASD, we set the contrastive strength to $\alpha = 1.0$ and the APC threshold to $\beta = 0.1$, keeping them consistent with the default settings of VCD~\cite{vcd}. Furthermore, the confidence gating threshold is set to $\tau = 0.5$, and the attention sparsity ratio is defined as $k = 0.1$, which selects the top $\lceil 0.1N \rceil$ visual tokens from $N$ visual tokens at each triggered step. The Gaussian perturbation scale $\sigma=1$ is fixed across all experiments rather than tuned per benchmark or backbone.

\textbf{Evaluation Benchmarks.} To evaluate the effectiveness of CHASD in hallucination mitigation, we employ five diverse benchmarks:
\begin{itemize}
    \item \textbf{POPE}~\cite{pope}: Formulates hallucination evaluation as a binary QA task across three different data distributions, serving as a core benchmark for object hallucination. In addition, POPE uses image-label pairs from three different large-scale datasets to ensure a comprehensive and robust evaluation across diverse visual contexts.
 \item \textbf{AMBER}~\cite{amber}: A multi-dimensional framework that systematically assesses overall reliability by providing a unified metric that aggregates performance across decoupled generative (image captioning) and discriminative (existence-based probing) tasks.
    \item \textbf{MME}~\cite{mme}: Performs fine-grained evaluation of hallucinations across 14 sub-tasks, from perception to cognition dimensions.
    \item \textbf{MMHal-Bench}~\cite{mmh}: Utilizes advanced LLMs, such as GPT-4o~\cite{gpt4o}, as evaluators to subjectively score the consistency of long-form generated responses, focusing on the balance between the richness of information and factual accuracy in the model’s answers.
    \item \textbf{CHAIR}~\cite{chair}: Strictly quantifies the proportion of hallucinated objects among all mentioned objects in long-form generation tasks.
\end{itemize}

For a detailed description of the procedures for the above benchmarks and the corresponding methods for calculating the metrics, please refer to Appendix~\ref{sec:benchmark}.

\textbf{Baselines.} We evaluate the performance of CHASD by comparing it against a suite of strong training-free strategies. These include \textbf{VCD}~\cite{vcd}, the pioneering work that introduced multi-modal contrastive decoding, as well as more recent attention-guided approaches such as \textbf{AvisC}~\cite{avisc} and \textbf{SID}~\cite{sid}. These comparisons are designed to evaluate whether CHASD's dynamic and localized intervention mechanism improves hallucination-related metrics under comparable inference-time settings.

\subsection{Experimental Results}

\textbf{Performance on POPE.} Table~\ref{tb:pope} summarizes the performance of various backbone models on the POPE benchmark. All values reported are the averages across MSCOCO~\cite{coco}, A-OKVQA~\cite{aokvqa}, and GQA~\cite{gqa} datasets. Empirical results show that, while CHASD achieves an F1 score comparable to the best baseline under the \textit{Random} setting with LLaVA-1.5, it obtains the best results across most backbone models and query configurations. Notably, when utilizing InstructBLIP as the backbone, CHASD achieves an absolute accuracy improvement of 1.39\% over the previous best method under the most challenging \textit{Adversarial} setting. This gain suggests that attention-guided local perturbation helps reduce over-extrapolation from misaligned or insufficient visual evidence.
\input{Tables/pope}

\textbf{Performance on AMBER.} Unlike single-task benchmarks, AMBER introduces a systematic evaluation of generated text content. As shown in Table~\ref{amber}, CHASD yields the best results on several core metrics, including CHAIR and F1-score, across both generative and discriminative tasks. Overall, CHASD attains the highest AMBER score. Specifically, compared to the best-performing baseline, CHASD achieves performance boosts of 1.5\% and 0.45\% on LLaVA-1.5 and InstructBLIP, respectively, indicating improved reliability across AMBER's mixed evaluation settings.
\input{Tables/amber}

\textbf{Performance on MME.} The MME benchmark provides a fine-grained exploration of models' perceptual capabilities, and the specific results are shown in Table~\ref{tb:mme}. CHASD achieves the highest total scores on both LLaVA-1.5 (592.35) and InstructBLIP (471.72) while exhibiting an exceptionally balanced performance distribution. Although certain baseline methods occasionally show localized advantages in specific sub-tasks—such as the performance of SID in the counting task under LLaVA—they suffer from significant performance degradation when transferred across architectures or tasks. In contrast, CHASD maintains a high level of stability across all evaluation dimensions without obvious perceptual bottlenecks. MME-Fullset results are deferred to Appendix~\ref{sec:more_results}.
\input{Tables/MME}

\textbf{Results on MMHal-Bench and CHAIR.} As illustrated in Table~\ref{tb:mmh}, CHASD improves several long-form hallucination metrics. On LLaVA-1.5, CHASD improves the MMHal-Bench score from 1.70 to 1.82 and reduces the hallucination rate to 0.70. Regarding the CHAIR metric, which directly quantifies object hallucination density, CHASD shows strong results on InstructBLIP: reducing $\text{CHAIR}_s$ from 52.2 to 44.0, and further reducing $\text{CHAIR}_i$ from 14.2 to 11.2. These results indicate that CHASD can improve visual faithfulness during long-form generation.

\input{Tables/MMH}

\textbf{Case Study on LLaVA-Bench.} Figure~\ref{fig:llavabench} intuitively visualizes text generation examples from different methods. From the red highlights (hallucinated segments), it is evident that Greedy decoding is highly susceptible to hallucinating object identities, spatial locations, and quantities. While VCD partially alleviates this issue, CHASD can most precisely anchor genuine visual evidence, significantly suppressing hallucination generation. For instance, in the right-side example, only CHASD correctly identifies the fruit as a mangosteen.

\input{Figures/llavabench}

\subsection{Ablation Study}

In this section, we first evaluate the computational efficiency of CHASD by comparing the inference latency and GPU memory footprint across various methods. Subsequently, we conduct a sensitivity analysis on the key hyperparameters, namely the attention sparsity $k$ and the confidence gating threshold $\tau$, to investigate their impact on the balance between performance and overhead. All of the above experiments were conducted on POPE benchmark (LLaVA-1.5, COCO, \textit{Adversarial}).

\begin{wraptable}{r}{0.55\textwidth}
    
    \centering
    \small
    \setlength{\tabcolsep}{4pt}
    \renewcommand{\arraystretch}{1.0}
    \caption{Comparison of different decoding methods on accuracy, runtime, and memory usage.}
    \label{tab:efficiency}
\begin{tabular}{lccc}
\toprule
\textbf{Method} & \textbf{Time (s)\down} & \textbf{Memory (MB)}\down & \textbf{Accuracy}\up \\
\midrule
Greedy  & 564  & 14328 & 0.7665 \\
VCD~\cite{vcd}     & 1073 & 14822 & 0.7645 \\
AVISC~\cite{avisc}   & 1170 & 16698 & 0.7798 \\
SID~\cite{sid}     & 1221 & 16628 & 0.8040 \\
\textbf{CHASD} & 1069 & 15594 & \textbf{0.8319} \\
\bottomrule
\end{tabular}

\end{wraptable}

\textbf{Efficiency Comparison.} In Table~\ref{tab:efficiency}, we compared the runtime and resource consumption of CHASD with other CD methods, and also reported their respective accuracy. Specifically, while achieving the best accuracy of 83.19\%, CHASD's inference latency was roughly on par with that of VCD. Compared to recently proposed attention-guided methods, our approach achieves a 9\% to 12\% acceleration in inference and saves approximately 1.0 GB of GPU memory usage. This efficiency advantage mainly stems from the confidence gate, which avoids negative-branch forward passes for high-confidence next-token distributions.

\input{Tables/ablation}

\textbf{Impact of Attention Sparsity $k$.} As shown in Figure~\ref{fig:k}, as $k$ decreases and the perturbation becomes more focused, the model performance consistently improves, peaking at $k=0.1$. This confirms that precisely perturbing the most salient visual tokens is significantly more effective than global blurring in exposing and penalizing spurious confidence induced by language priors.

\textbf{Trade-off between Gating Threshold $\tau$ and Efficiency.} Figure~\ref{fig:tau} illustrates the trade-off between predictive performance and computational latency with respect to $\tau$. While $\tau=1.0$ (universal calibration) approaches the performance upper bound, it requires more negative-branch forward passes. We observe a stable performance plateau for $\tau \in [0.5, 0.9]$, alongside an efficiency inflection point at $\tau=0.5$. Considering this trade-off, we adopt $\tau=0.5$ as the default configuration. Under this setting, CHASD saves over 50\% of inference time while incurring a 0.18\% drop in accuracy. This result supports the use of uncertainty-based adaptive gating to reduce redundant computation while preserving generation quality.

%% file: Tables/pope.tex
\begin{table*}[t]
\centering
\small

\caption{\textbf{Comparison results from different CD methods on POPE benchmark}. The results reported in the table are the average results across three datasets.}
\label{tb:pope}
\resizebox{\textwidth}{!}{
\begin{tabular}{llcccccc}
\toprule
\multicolumn{2}{c}{\textbf{Setting}} 
& \multicolumn{2}{c}{\textit{Random}} 
& \multicolumn{2}{c}{\textit{Popular}} 
& \multicolumn{2}{c}{\textit{Adversarial}} \\
\cmidrule(lr){3-4} \cmidrule(lr){5-6} \cmidrule(lr){7-8}
\textbf{Model} & \textbf{Method} 
& Acc & F1
& Acc & F1 
& Acc & F1 \\
\midrule

\multirow{5}{*}{LLaVA-1.5~\cite{llava}}

& Greedy     & 81.19 & 80.15 & 76.81 & 76.68 & 72.97 & 73.57 \\
& VCD~\cite{vcd}      & 82.51 & 83.96 & 77.12 & 75.28 & 71.96 & 76.54 \\
& AvisC~\cite{avisc}    & 85.21 & 86.22 & 78.65 & 81.37 & 71.64 & 76.66 \\
& SID~\cite{sid}      & \cellcolor{cyan!5}88.11 & \cellcolor{cyan!17}88.48& \cellcolor{cyan!5}81.85 & \cellcolor{cyan!5}83.52 & \cellcolor{cyan!5}75.56 & \cellcolor{cyan!5}78.96 \\
& \textbf{Ours(CHASD)} 
& \cellcolor{cyan!17}88.58
& \cellcolor{cyan!5}87.93 
& \cellcolor{cyan!17}84.71
& \cellcolor{cyan!17}84.57
& \cellcolor{cyan!17}80.55
& \cellcolor{cyan!17}81.20 \\

\midrule

\multirow{5}{*}{InstructBLIP~\cite{instructblip}}

& Greedy     & 81.80 & 81.33 & 77.37 & 77.87& 74.48 & 75.64 \\
& VCD~\cite{vcd}      & 84.50 & 84.11 & \cellcolor{cyan!5}81.77 & 81.88 & \cellcolor{cyan!5}78.20 & \cellcolor{cyan!5}78.98 \\
& AvisC~\cite{avisc}    & 86.14 & 86.16 & 79.76 & 80.96 & 74.73 & 77.47 \\
& SID~\cite{sid}      & \cellcolor{cyan!5}87.23 & \cellcolor{cyan!5}86.90 & 81.00 & \cellcolor{cyan!5}82.30 & 74.45 & 77.57 \\
& \textbf{Ours(CHASD)} 
& \cellcolor{cyan!17}88.02
& \cellcolor{cyan!17}87.23
& \cellcolor{cyan!17}83.05
& \cellcolor{cyan!17}82.91
& \cellcolor{cyan!17}79.59 
& \cellcolor{cyan!17}80.15 \\

\bottomrule
\end{tabular}
}

\end{table*}

%% file: Tables/amber.tex
\begin{table*}[t]
\centering
\small
\caption{\textbf{AMBER results for different CD methods and backbones.} CHASD achieved the highest scores on multiple sub-task metrics as well as the overall composite metric}
\label{amber}
\resizebox{\textwidth}{!}{
\begin{tabular}{ll cccc cccc c}
\toprule

\multirow{2}{*}{\textbf{Model}} 
& \multirow{2}{*}{\textbf{Method}}
& \multicolumn{4}{c}{\textbf{Generative}} 
& \multicolumn{4}{c}{\textbf{Discriminative}} 
& \multirow{2}{*}{\textbf{AMBER}\up} \\

\cmidrule(lr){3-6} \cmidrule(lr){7-10}

& 
& CHAIR\down & Cover\up & Hal\down & Cog\down
& Acc.\up & Prec.\up & Rec.\up & F1\up
& \\

\midrule

\multirow{5}{*}{\rotatebox{90}{LLaVA-1.5}}
& Greedy 
& 8.90 & 45.00 & 31.50 &2.20 
& 69.90& 83.70 &\cellcolor{cyan!5} 68.10 & 75.10 
& 83.10 \\

& VCD~\cite{vcd}
& 6.60 & \cellcolor{cyan!17}49.30 &28.50 &\cellcolor{cyan!5} 2.00 
& 67.80 & \cellcolor{cyan!5}93.30 &55.60 &69.70 
& 81.55 \\

& AvisC ~\cite{avisc}
& 7.20 & 45.70 & 29.10 &2.30 
& 70.90 & 85.10 & 68.00 & 75.60
& 84.20 \\

& SID ~\cite{sid}
& \cellcolor{cyan!5}6.20 &\cellcolor{cyan!5} 47.50& \cellcolor{cyan!17}19.40 & \cellcolor{cyan!17}1.60 
&\cellcolor{cyan!5} 74.90 & \cellcolor{cyan!17}93.60& 66.80 & \cellcolor{cyan!5}78.00 
& \cellcolor{cyan!5}85.90 \\

& \textbf{Ours(CHASD)} 
& \cellcolor{cyan!17}6.10 & 45.90 & \cellcolor{cyan!5}24.70 & 2.20
& \cellcolor{cyan!17}76.60 & 88.60 & \cellcolor{cyan!17}74.50 & \cellcolor{cyan!17}80.90
& \cellcolor{cyan!17}87.40 \\

\midrule

\multirow{5}{*}{\rotatebox{90}{InstructBLIP}}
& Greedy 
& 9.80 & 45.90 & 35.20 &2.70 
& 68.30& 78.20 & 71.50 & 74.70 
& 82.45 \\

& VCD ~\cite{vcd}
& 5.90 &\cellcolor{cyan!17}50.00 &26.90 & 2.10 
& 75.40 & \cellcolor{cyan!17}84.10 &76.70 &80.20 
& 87.15 \\

& AvisC ~\cite{avisc}
& 7.50 & 47.00 & 29.10 &2.50 
& 73.00 & 81.50 & 76.80 & 79.10
& 85.80 \\

& SID ~\cite{sid}
& \cellcolor{cyan!5}5.60 & \cellcolor{cyan!5}48.00& \cellcolor{cyan!17}22.70 &\cellcolor{cyan!5}2.00 
& \cellcolor{cyan!5}75.80 & 83.20& \cellcolor{cyan!5}79.60 & \cellcolor{cyan!5}81.40 
& \cellcolor{cyan!5}87.90 \\

& \textbf{Ours(CHASD)} 
&\cellcolor{cyan!17}5.30 & 47.30 &\cellcolor{cyan!5} 22.90 & \cellcolor{cyan!17}1.80
& \cellcolor{cyan!17}76.90 & \cellcolor{cyan!5}83.70 & \cellcolor{cyan!17}80.40 & \cellcolor{cyan!17}82.00
& \cellcolor{cyan!17}88.35 \\

\bottomrule
\end{tabular}
}
\vspace{-1em}
\end{table*}

%% file: Tables/MME.tex
\begin{table}[t]
\centering

\caption{\textbf{MME benchmark} (Object-level and Attribute-level) results.}
\label{tb:mme}
\resizebox{\textwidth}{!}{
\begin{tabular}{l l c c c c c}
\toprule
\multirow{2}{*}{Model} & \multirow{2}{*}{Method} 
& \multicolumn{2}{c}{Object-level} 
& \multicolumn{2}{c}{Attribute-level} 
& \multirow{2}{*}{Total Score \up} \\
\cmidrule(lr){3-4} \cmidrule(lr){5-6}
& & Existence \up & Count \up
& Position \up & Color \up & \\
\midrule

\multirow{5}{*}{LLaVA-1.5~\cite{llava}}
& Greedy      & 173.57 & 125.82 & 118.33 & \cellcolor{cyan!5}114.02 & 531.74 \\
& VCD ~\cite{vcd}      & 172.14 & 120.14 & 115.00 & 108.96 & 516.24 \\
& AvisC~\cite{avisc}     & \cellcolor{cyan!17}189.29 & 115.76 & 119.13 & \cellcolor{cyan!17}120.00 & 544.18 \\
& SID ~\cite{sid}     & \cellcolor{cyan!5}185.00 & \cellcolor{cyan!5}143.33 &\cellcolor{cyan!5} 133.33 & \cellcolor{cyan!17}120.00 & \cellcolor{cyan!5}581.66 \\
& \textbf{Ours(CHASD)}    & \cellcolor{cyan!5}185.00 & \cellcolor{cyan!17}148.33 & \cellcolor{cyan!17}145.00 & \cellcolor{cyan!5}114.02 & \cellcolor{cyan!17}592.35 \\

\midrule

\multirow{5}{*}{InstructBLIP~\cite{instructblip}}
& Greedy      & 155.00 & \cellcolor{cyan!5}65.00  & 55.00  &\cellcolor{cyan!5} 138.33 & 413.33 \\
& VCD ~\cite{vcd}      & 183.00 & 60.00  & 50.00  & 125.00 & 418.00 \\
& AvisC ~\cite{avisc}    & \cellcolor{cyan!5}185.00 & \cellcolor{cyan!17}80.00  & \cellcolor{cyan!17}63.33  & 120.00 &\cellcolor{cyan!5} 448.33 \\
& SID   ~\cite{sid}    & \cellcolor{cyan!5}185.00 & 60.00  & 50.00  & 120.00 & 415.00 \\
& \textbf{Ours(CHASD)}    & \cellcolor{cyan!17}190.00 & \cellcolor{cyan!17}80.00  & \cellcolor{cyan!5}58.39  & \cellcolor{cyan!17}143.33 & \cellcolor{cyan!17}471.72 \\

\bottomrule
\end{tabular}}
\end{table}

%% file: Tables/MMH.tex
\begin{table}[t]
\centering

\caption{\textbf{MMHal-Bench and Object-Hallucination} results.}
\label{tb:mmh}

\begin{tabular}{l l c c c c}
\toprule
\multirow{2}{*}{Model} & \multirow{2}{*}{Method} 
& \multicolumn{2}{c}{MMHal-Bench} 
& \multicolumn{2}{c}{Object-Hallucination} \\
\cmidrule(lr){3-4} \cmidrule(lr){5-6}
& & Score \up & HalRate \down 
& CHAIR$_s$ \down & CHAIR$_i$ \down \\
\midrule

\multirow{5}{*}{LLaVA-1.5~\cite{llava}}
& Greedy          & \cellcolor{cyan!5}1.70 &\cellcolor{cyan!5} 0.71 & 49.6 & \cellcolor{cyan!5}14.0 \\
& VCD~\cite{vcd}           & 1.31 & 0.81 & 50.6 & 15.7 \\
& AvisC~\cite{avisc}          & 1.69 & 0.73 & 59.4 & 17.5 \\
& SID ~\cite{sid} & 1.69   & 0.72   & \cellcolor{cyan!5}49.0& \cellcolor{cyan!17}13.5 \\
& \textbf{Ours(CHASD)}          &\cellcolor{cyan!17} 1.82 &\cellcolor{cyan!17}0.70 &\cellcolor{cyan!17}47.0 & 14.2 \\

\midrule

\multirow{5}{*}{InstructBLIP~\cite{instructblip}}
& Greedy          & 1.55 &\cellcolor{cyan!17} 0.70 & 52.2 & 21.6 \\
& VCD~\cite{vcd}           & 1.48 & 0.74& 51.8 &\cellcolor{cyan!5}14.2 \\
& AvisC~\cite{avisc}          & \cellcolor{cyan!5}1.56 & \cellcolor{cyan!5}0.71 & 59.2 & 24.4 \\
& SID~\cite{sid}  & 1.54   &\cellcolor{cyan!17}0.70   &\cellcolor{cyan!5} 50.0 & 21.4 \\
& \textbf{Ours(CHASD)}          & \cellcolor{cyan!17}1.57 & \cellcolor{cyan!5}0.71 & \cellcolor{cyan!17}44.0 & \cellcolor{cyan!17}11.2 \\

\bottomrule
\end{tabular}
\end{table}

%% file: Figures/llavabench.tex
\newpage
\begin{figure}[h]
\centering

\begin{minipage}[t]{0.495\linewidth}
\centering
\includegraphics[width=\linewidth]{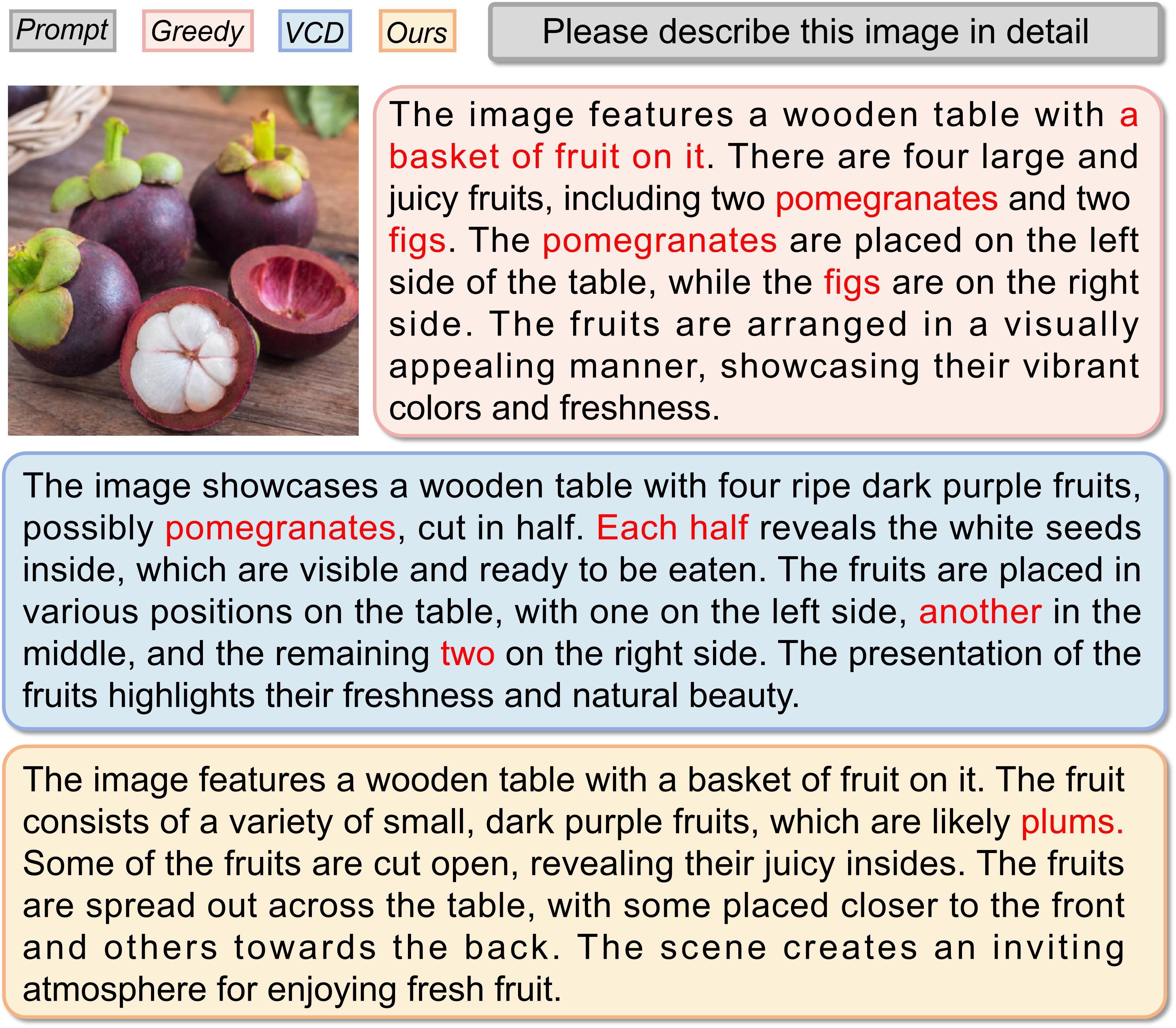}
\end{minipage}
\hfill
\begin{minipage}[t]{0.495\linewidth}
\centering
\includegraphics[width=\linewidth]{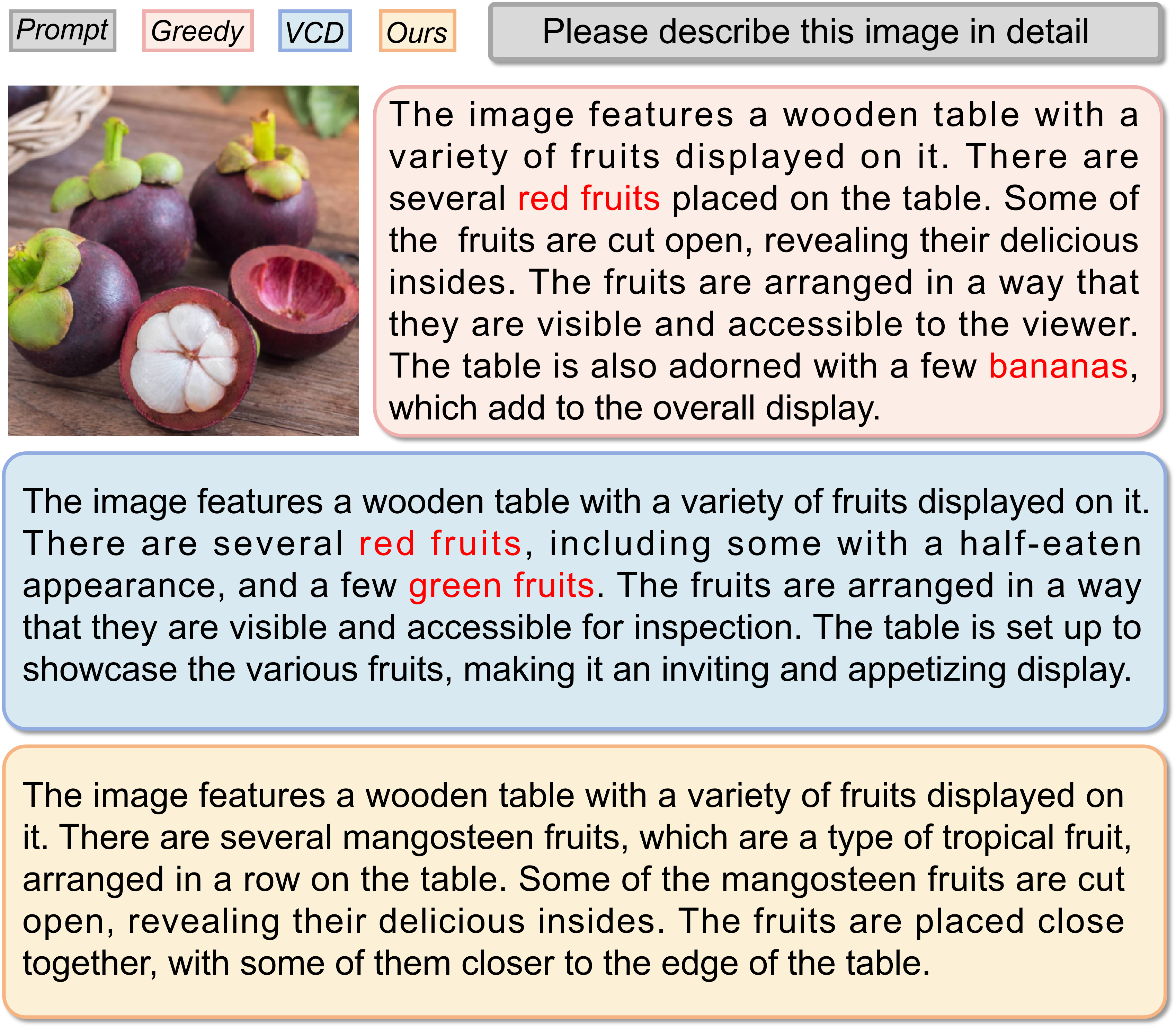}
\end{minipage}

\caption{\textbf{Comparison of image descriptions} generated by different methods on LLaVA-Bench (Left: LLaVA; Right: InstructBLIP). Hallucinated content is highlighted in \textcolor{red}{red}.}
\label{fig:llavabench}
\end{figure}

%% file: Tables/ablation.tex
\begin{figure}[t]
\centering

\begin{minipage}[t]{0.45\linewidth}
\centering

\captionsetup{type=figure}

\vspace{2pt}

\includegraphics[width=1\linewidth]{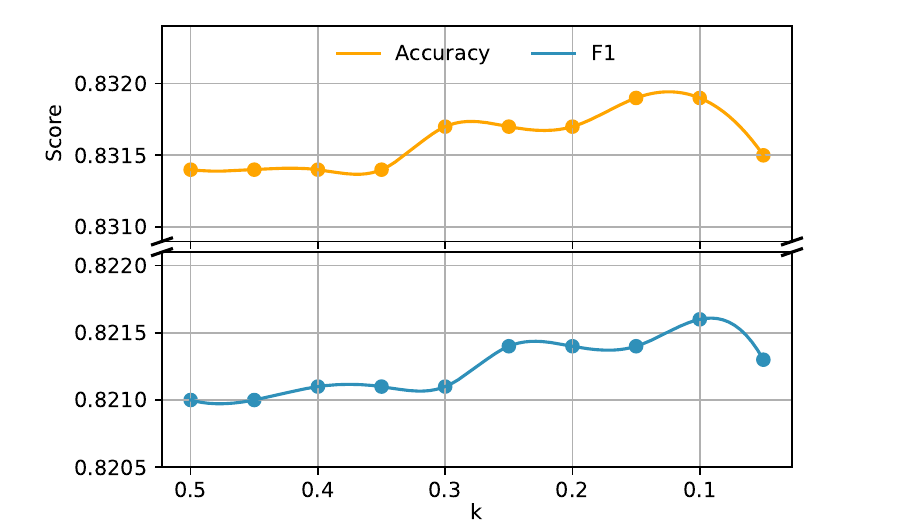}
\caption{Sensitivity of the hyperparameter for the \textbf{attention sparsity} $k$}
\label{fig:k}
\end{minipage}%
\hspace{0.05\linewidth}%
\begin{minipage}[t]{0.45\linewidth}
\centering

\captionsetup{type=figure}

\vspace{2pt}

\includegraphics[width=1\linewidth]{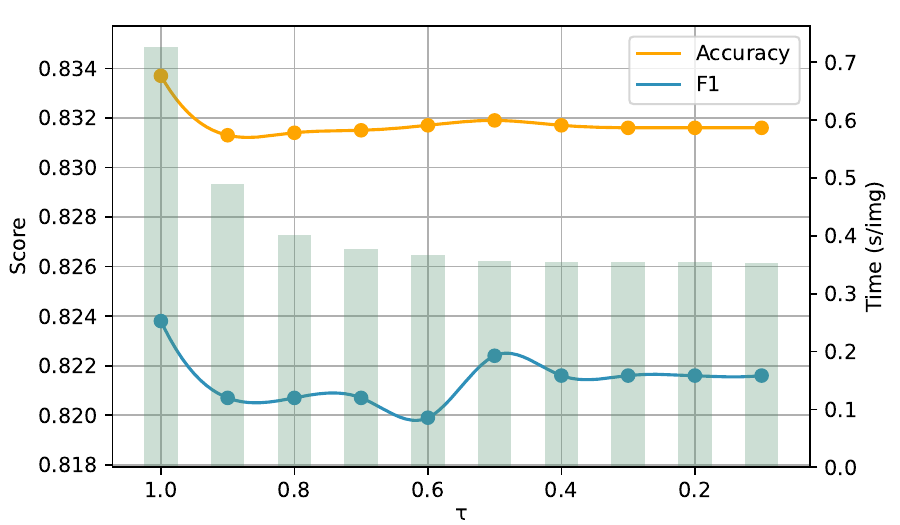}
\caption{Sensitivity of the hyperparameter for the \textbf{confidence gating threshold} $\tau$.}
\label{fig:tau}
\end{minipage}%
\end{figure}

%% file: Sections/5.conclusion.tex
\section{Conclusion}

In this paper, we propose \textbf{CHASD}, a training-free inference framework designed to mitigate object hallucinations in LVLMs. Addressing the \textbf{spatiotemporal heterogeneity} of hallucination risks, we introduce a \textbf{spatiotemporal coupled constraint} to achieve \textbf{"calibration on demand,"} reducing the need for persistent, global perturbations. By combining \textbf{uncertainty-driven confidence gating} with \textbf{attention-guided localized perturbation}, CHASD applies contrastive calibration only to selected low-confidence decoding steps and perturbs the currently salient visual tokens for the negative branch. Evaluations across POPE, AMBER, MME, MMHal-Bench, and CHAIR show that CHASD improves hallucination-related metrics over strong training-free baselines while reducing unnecessary negative-branch computation. A remaining limitation is that the confidence gate may bypass high-confidence hallucinated tokens, suggesting that future work could incorporate richer uncertainty signals beyond maximum token probability.

%% file: Sections/appendix.tex
\newpage
\section*{Appendix Overview}

This appendix provides additional details and analyses to complement the main paper. It is organized as follows:

\begin{itemize}
    \item \textbf{Section~\ref{sec:impact}.~Social Impact.} We discuss the ethical implications of the \textbf{CHASD} framework, focusing on its contribution to AI reliability.
    
    \item \textbf{Section~\ref{sec:limitations}.~Limitations.} We acknowledge and analyze the potential risks of model over-confidence and the generalization of hyperparameters across different model scales.
    
    \item \textbf{Section~\ref{sec: Complexity}.~Comprehensive Complexity Analysis.} We provide a theoretical analysis of the time and space complexity of \textbf{CHASD}, focusing on the effect of confidence-gated negative-branch computation.
    
    \item \textbf{Section~\ref{sec:benchmark}.~Detailed Benchmark Specifications and Metric Calculations.} We detail the taxonomy and mathematical formulations for POPE, AMBER, MME, and MMHal-Bench.
    
    \item \textbf{Section~\ref{sec:more_results}.~More Results.} We present the MME-Fullset performance and extend our evaluation to more recent and larger-scale backbones, including Qwen2.5-VL~\cite{Qwen2.5-VL} and LLaVA-1.5-13B~\cite{llava}.
    
    \item \textbf{Section~\ref{sec:visualization}.~Visualizations.} We provide qualitative comparisons of image descriptions to highlight the efficacy of our method in mitigating hallucinations.
    
    \item \textbf{Section~\ref{sec:prompts}.~Prompts for GPT-4o~\cite{gpt4o}.} We disclose the full prompt templates used to enable GPT-4o to act as a judge during the MMHal-Bench evaluation.
\end{itemize}

\section{Social Impact}
\label{sec:impact}

The proposed CHASD framework contributes to the development of trustworthy and sustainable AI by effectively mitigating object hallucinations while maintaining high inference efficiency. By enabling "\textbf{calibration on demand}," our work enhances the reliability of LVLMs in critical, high-stakes applications—such as automated medical diagnosis and assistive technologies—while reducing the carbon footprint typically associated with dual-branch contrastive inference. While CHASD promotes more "honest" AI outputs, it remains susceptible to biases inherited from the backbone models; hence, we emphasize the importance of pairing this framework with robust ethical safeguards and bias-detection tools to ensure its responsible deployment and minimize the risk of generating convincing misinformation.

\section{Limitations}
\label{sec:limitations}
While CHASD establishes a robust and efficient paradigm for mitigating object hallucinations, we acknowledge several limitations that warrant further investigation.

 \textbf{Risks of Model Over-confidence.} CHASD relies on the uncertainty-driven gating mechanism to determine when to intervene. However, LVLMs occasionally exhibit overconfidence—assigning high probability ($p \ge \tau$) to hallucinated tokens due to overwhelming language priors. In such rare cases, CHASD might bypass the calibration process (Stage I in Figure~\ref{fig:pipeline}), failing to rectify the error. Future work could explore integrating multi-dimensional uncertainty metrics (e.g., entropy or semantic consistency) to enhance the sensitivity of the gating trigger.

\textbf{Generalization of Hyperparameters.} Although we provide sensitivity analyses for the threshold $\tau$ and sparsity $k$, the optimal values may vary across different model scales (e.g., 7B vs. 70B) or vision-language alignment strategies (e.g., CLIP-based vs. SigLIP-based encoders). Developing an automated, model-agnostic parameter tuning mechanism would be a promising direction for enhancing the plug-and-play capability of CHASD.

\section{Comprehensive Complexity Analysis}
\label{sec: Complexity}
In this section, we provide a rigorous theoretical analysis of the computational and memory efficiency of \textbf{CHASD} compared to vanilla decoding and standard contrastive decoding methods.

\subsection{Time Complexity}
Let $L$ be the sequence length and $T_{fwd}$ be the time for a single model forward pass.
\begin{itemize}
    \item \textbf{Vanilla Decoding:} Each token is generated with one forward pass, resulting in $O(L \cdot T_{fwd})$.
    \item \textbf{Standard CD} (e.g., VCD)~\textbf{:} Requires dual forward passes (original and negative) at every step, yielding a constant complexity of $O(2L \cdot T_{fwd})$.
    \item \textbf{CHASD (Ours):} By introducing the \textit{confidence gate} $\tau$, the negative branch is only activated when the maximum next-token probability $p < \tau$. Let $\theta \in [0, 1]$ denote the \textbf{trigger rate} (the ratio of tokens requiring calibration). The expected number of forward passes is $(1 + \theta)L$, and the expected time complexity is $O[(1 + \theta) L \cdot T_{fwd}]$.

\end{itemize}

\subsection{Space Complexity and Memory Footprint}
\begin{itemize}
    \item \textbf{Weight Redundancy:} As a training-free framework, CHASD requires zero additional trainable parameters, maintaining $O(1)$ parameter growth.
    \item \textbf{Negative-Branch Cache Usage:} Standard CD evaluates the negative branch at every decoding step. CHASD invokes this branch only at triggered steps, so auxiliary cache usage and related activation memory are proportional to the trigger rate in implementations that allocate the negative branch on demand.
    \item \textbf{Localized Masking:} CHASD's \textit{attention-guided localized perturbation} selects $m=\lceil kN \rceil$ visual tokens for corruption while preserving the full visual token sequence as input. The memory overhead for the dynamic mask $\mathcal{M}_t$ is $O(N)$ bits, which is negligible compared to model activations.
\end{itemize}

\section{Detailed Benchmark Specifications and Metric Calculations}
\label{sec:benchmark}
\subsection{Benchmark Specifications and Taxonomy}
\begin{enumerate}
    \item \textbf{POPE~\cite{pope}:} To mitigate the bias of simple "Yes" answers, POPE employs three progressively difficult sampling settings:
    \begin{itemize}
        \item \textit{Random:} Negative objects are randomly sampled from those not present in the image.
        \item \textit{Popular:} Negative objects are selected from the most frequent categories in the dataset (e.g., "person," "chair").
        \item \textit{Adversarial:} Negative objects are selected from those that frequently co-occur with ground-truth objects but are absent in the current image (e.g., querying "sink" when "toilet" is present).
    \end{itemize}
     \item \textbf{AMBER~\cite{amber}:} AMBER spans both \textit{Discriminative} (Yes/No questions based on POPE-style sampling) and \textit{Generative} (Image Captioning) tasks to provide a multi-faceted assessment.
   \item \textbf{MME~\cite{mme}:} A comprehensive benchmark designed to evaluate both perception and cognition capabilities. 
\begin{itemize}
    \item The \textit{Perception} track comprises 10 sub-tasks: Existence, Count, Color, Position, Celebrity, Landmark, Artwork, Poster, Movie, and Design, focusing on basic visual recognition.
    \item The \textit{Cognition} track consists of the remaining 4 sub-tasks: Commonsense Reasoning, Numerical Computing, Code Reasoning, and Text Translation, which require higher-level logical deduction.
\end{itemize}

    \item \textbf{MMHal-Bench~\cite{mmh}:} This benchmark evaluates hallucinatory responses across 8 complex reasoning dimensions: Attribute, Comparison, Counting, Existence, Localization, Relation, Scene, and Sport.

\end{enumerate}

\subsection{Precise Metric Calculations}
\begin{enumerate}
    \item \textbf{POPE Metrics:} We report Accuracy ($\mathrm{Acc}$) and F1-score ($\mathrm{F1}$) to analyze the trade-off between sensitivity and specificity:
 \begin{equation}
    \mathrm{Acc} = \frac{TP + TN}{TP + TN + FP + FN}, \quad \mathrm{F1} = \frac{2 TP}{2 TP + FP + FN}
    \end{equation}

    \item \textbf{AMBER Score Calculation:} The final AMBER score is computed as the arithmetic mean of the discriminative F1-score and the generative fidelity (represented by $100 - \text{CHAIR}_i$), providing a balanced metric for both task types:
    \begin{equation}
    \text{Score}_{AMBER} = \frac{(100 - \text{CHAIR}_i) + \text{F1}}{2}
    \end{equation}
    Where $\text{CHAIR}_i$ denotes the instance-level hallucination rate in generative tasks and $\text{F1}_{pope}$ is the F1-score in discriminative tasks.

    \item \textbf{MME Scoring:} For each category $c$, the score is the sum of raw accuracy ($Acc_c$) and balanced accuracy ($Acc+_c$). Let $N_c$ be the number of image-question pairs in category $c$:
    \begin{equation}
    Acc_c = \frac{\sum \text{Correct Responses}}{N_c}, \quad \text{Score}_{MME} = \sum_{c=1}^{14} (Acc_c \times 100 + Acc+_c \times 100)
    \end{equation}
    The total maximum score for the Perception track is 2000.

    \item \textbf{MMHal-Bench Scoring:} The final informative-hallucination score ($S_{total}$) is the average across all 96 samples evaluated by an LLM-based judge:
    \begin{equation}
    S_{total} = \frac{1}{96} \sum_{j=1}^{96} \text{GPT-4-Judge}(\text{Response}_j)
    \end{equation}
\end{enumerate}


\section{More results}
\label{sec:more_results}
\subsection{MME-Fullset}
\begin{figure}[!htbp]
\centering
\includegraphics[width=\linewidth]{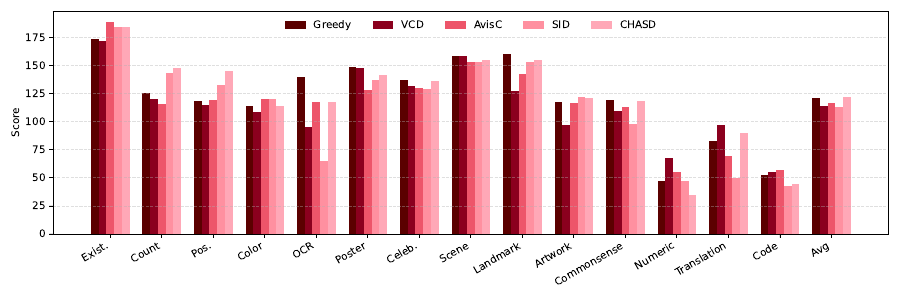}
\vspace{-1.5em}
\caption{MME-Fullset results for different methods (LLaVA-1.5 as backbone)}
\label{fig:llava_mmefull}
\end{figure}

\begin{figure}[!htbp]
\centering
\includegraphics[width=\linewidth]{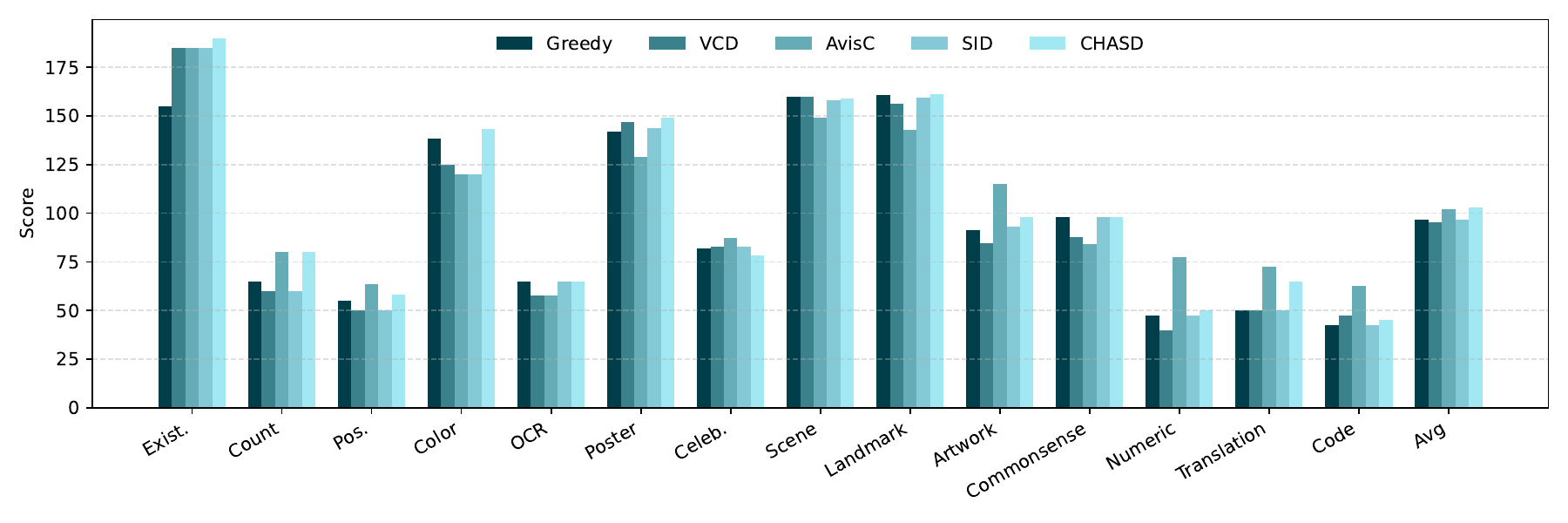}
\vspace{-1.5em}
\caption{MME-Fullset results for different methods (InstructBLIP as backbone)}
\label{fig:blipmmefull}
\end{figure}

Detailed subtask evaluations on the MME benchmark are presented here. Figures~\ref{fig:llava_mmefull} and~\ref{fig:blipmmefull} illustrate the performance profiles for LLaVA-1.5~\cite{llava} and InstructBLIP~\cite{instructblip}. While baseline methods like VCD~\cite{vcd} excel in particular areas such as numerical reasoning, they often fail to maintain a balanced improvement across all categories. In contrast, CHASD outperforms all counterparts in terms of the aggregate score on both architectures, validating its effectiveness as a comprehensive hallucination mitigation framework.

\subsection{Results on Other Backbones}
To further evaluate the versatility of CHASD, we extend our experiments to two additional backbones: the more recent Qwen2.5-VL~\cite{Qwen2.5-VL} and the larger-scale LLaVA-1.5-13B~\cite{llava}. As illustrated in Table~\ref{tb:pope_appx}, CHASD consistently delivers superior performance, underscoring its robust \textbf{generalization capabilities} across diverse model architectures.

\section{Visualizations}
\label{sec:visualization}
\begin{figure}[H]

\begin{minipage}[t]{0.495\linewidth}
\centering
\includegraphics[width=\linewidth]{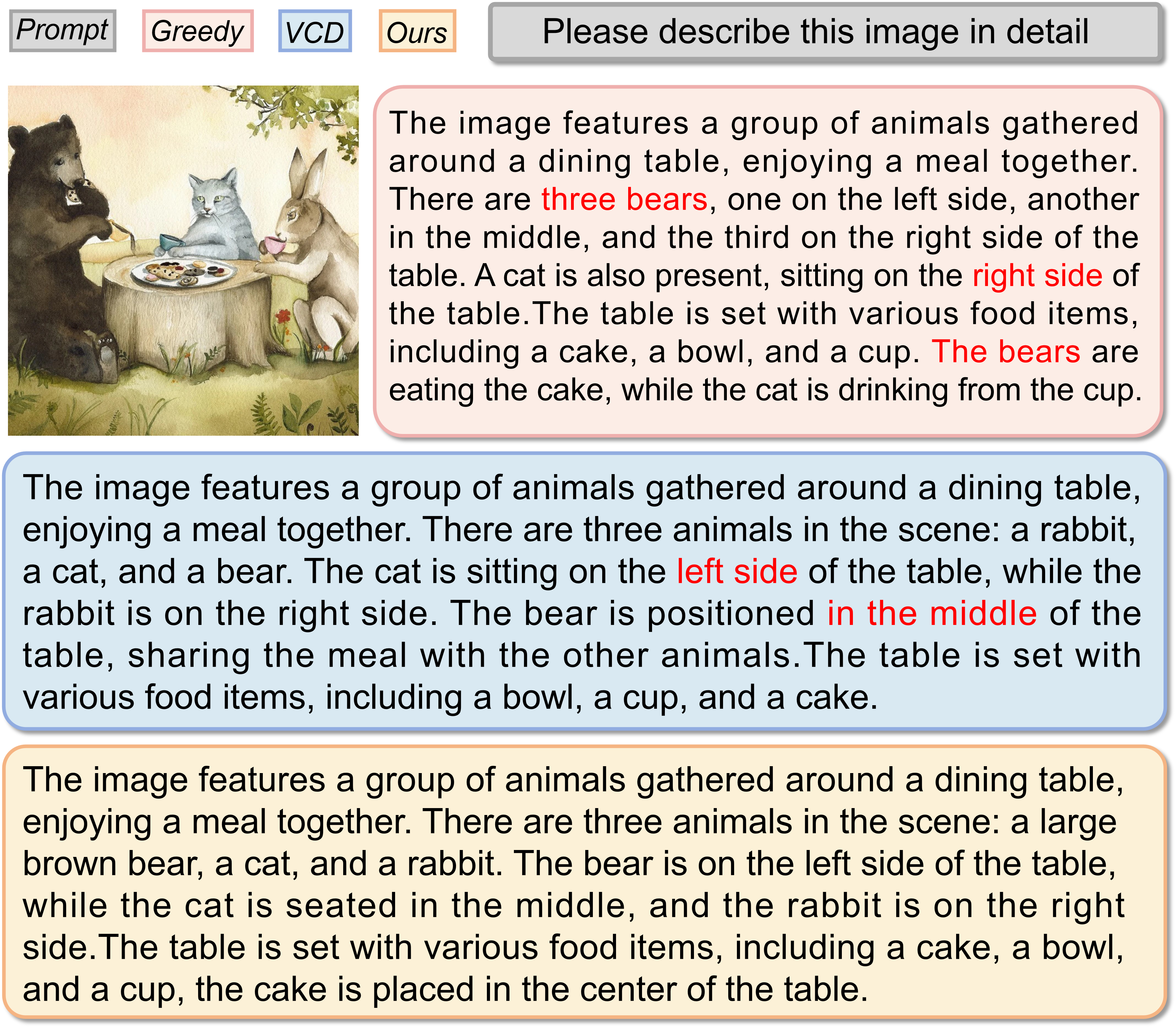}
\end{minipage}
\hfill
\begin{minipage}[t]{0.495\linewidth}
\centering
\includegraphics[width=\linewidth]{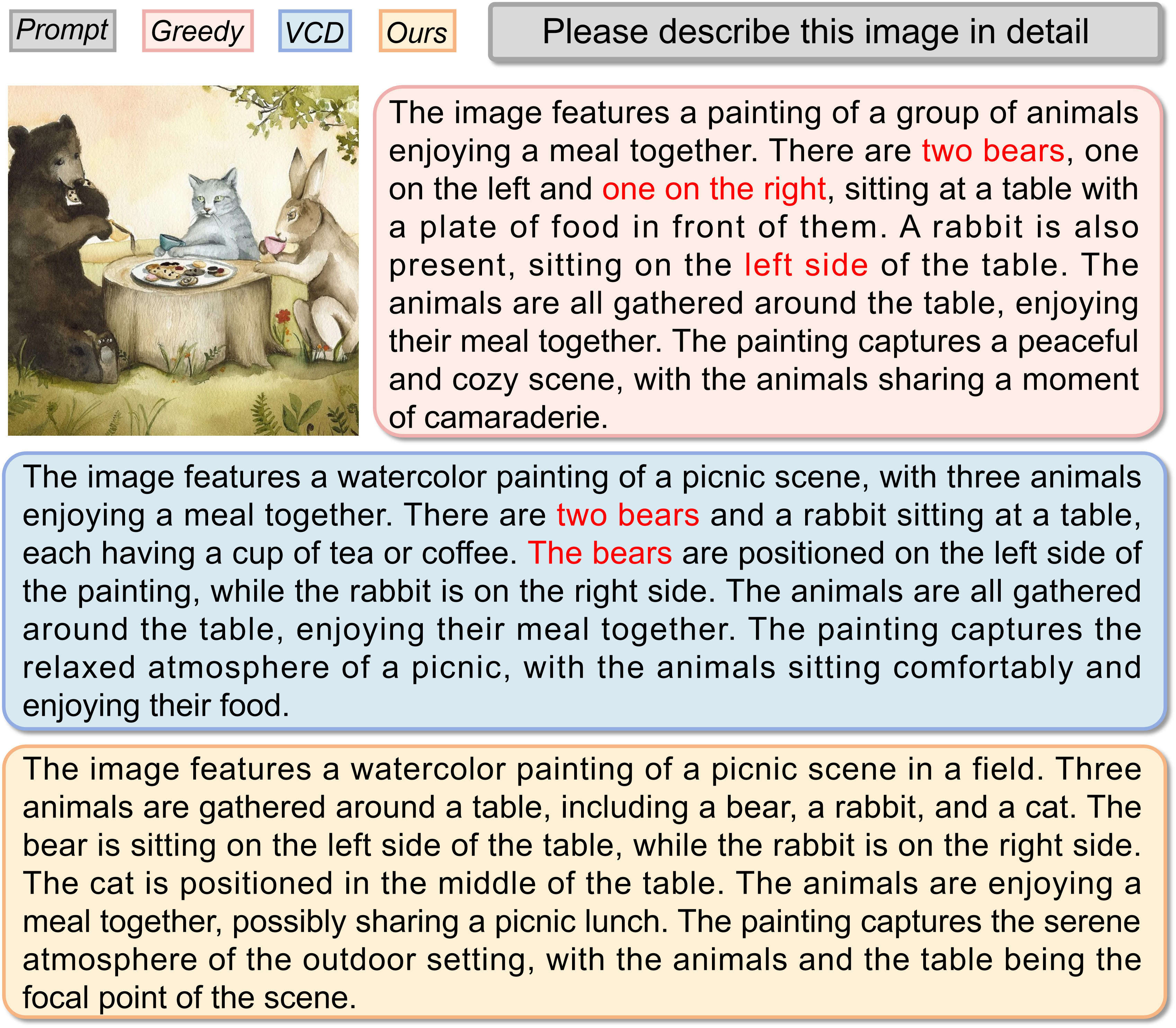}
\end{minipage}

\caption{\textbf{Comparison of image descriptions} generated by different methods on LLaVA-Bench (Left: LLaVA; Right: InstructBLIP). Hallucinated content is highlighted in \textcolor{red}{red}.}
\label{fig:more_vis}
\end{figure}

In Figure~\ref{fig:more_vis}, we provide further visualizations of hallucinations in the text responses generated by each model.

\input{Appendix/appx_tables/pope_more}

\section{Prompts for GPT-4o}
\label{sec:prompts}
During the MMHal-Benchmark evaluation, we used GPT-4o~\cite{gpt4o} to assist with scoring. Here, we will present the specific prompt we used to enable it to act as a judge.
\input{Appendix/appx_tables/prompts}

%% file: Appendix/appx_tables/pope_more.tex
\begin{table*}[!htbp]
\centering
\small
\vspace{1em}
\caption{\textbf{POPE benchmark results} for more backbones. The experiments in the table are conducted on the COCO dataset.}
\label{tb:pope_appx}
\resizebox{\textwidth}{!}{
\begin{tabular}{llcccccc}
\toprule
\multicolumn{2}{c}{\textbf{Setting}} 
& \multicolumn{2}{c}{\textit{Random}} 
& \multicolumn{2}{c}{\textit{Popular}} 
& \multicolumn{2}{c}{\textit{Adversarial}} \\
\cmidrule(lr){3-4} \cmidrule(lr){5-6} \cmidrule(lr){7-8}
\textbf{Model} & \textbf{Method} 
& Acc & F1
& Acc & F1 
& Acc & F1 \\
\midrule
\multirow{4}{*}{Qwen2.5-VL~\cite{Qwen2.5-VL}}

& Greedy     & 81.73 & 77.94 & 81.23 & 77.33& 80.83 & 77.08 \\
& VCD ~\cite{vcd}     & 82.67 & 79.23 & 82.97 & 79.89 & 81.77 & 78.35 \\
& AvisC~\cite{avisc}    & \cellcolor{cyan!5}85.63 &\cellcolor{cyan!5} 83.43 &\cellcolor{cyan!5} 84.57 & \cellcolor{cyan!5}82.36 & \cellcolor{cyan!5}83.90 &\cellcolor{cyan!5}   82.03 \\
& \textbf{Ours(CHASD)} &\cellcolor{cyan!17}86.37&\cellcolor{cyan!17} 84.31& \cellcolor{cyan!17}85.57& \cellcolor{cyan!17}83.54& \cellcolor{cyan!17}84.96 & \cellcolor{cyan!17}82.97 \\

\midrule

\multirow{4}{*}{LLaVA-1.5-13B~\cite{llava}}

& Greedy     & 79.33 & 79.63 & 76.30 & 77.32 & 73.44 & 75.12 \\
& VCD~\cite{vcd}      & 86.63 & 86.45 & 83.42 & 83.73 & \cellcolor{cyan!5}79.69 & 80.66 \\
& AvisC~\cite{avisc}    & \cellcolor{cyan!5}88.40 & \cellcolor{cyan!17}88.77 & \cellcolor{cyan!5}85.73 & \cellcolor{cyan!5}86.53 & 79.27 & \cellcolor{cyan!5}81.47 \\
& \textbf{Ours(CHASD)} 
& \cellcolor{cyan!17}88.89& \cellcolor{cyan!5}87.97 & \cellcolor{cyan!17}87.03& \cellcolor{cyan!17}86.75& \cellcolor{cyan!17}84.91& \cellcolor{cyan!17}84.35 \\

\bottomrule
\end{tabular}
}

\end{table*}

%% file: Appendix/appx_tables/prompts.tex
\lstset{
    basicstyle=\small\ttfamily,
    columns=flexible,
    breaklines=true, 
    frame=single,    
    backgroundcolor=\color{gray!10}
}

\begin{lstlisting}[]

Please act as an impartial and objective judge and evaluate the quality of the response provided by a Large Multimodal Model (LMM) to the user question. Your evaluation should be mainly based on whether the response is informative, and whether the response contains any hallucination. Hallucination, in this context, refers to a situation where the LMM generates a response that includes information not present or implied in the image or previous conversation. A hallucination could be a false claim about an object, action, emotion, or any other detail that is not grounded in the image.

For clarity, consider these examples:

### Example 1:
- Image Contents: Lynx, Cat
- Question: How many cats can you see in this image?
- Standard Human-Generated Answer: There are two cats visible in the image.
- LMM Response to Evaluate: There is one cat in the image.
- Analysis: The LMM's response is false because it claims there is only one cat in the image, when in fact there are two.
- Hallucination: Yes.

### Example 2:
- Image Contents: Tree, Plant, Panda, Animal
- Question: What can you see from this photo?
- Standard Human-Generated Answer: In this photo we can see branches. We can see a panda on a branch. The background is blurry.
- LMM Response to Evaluate: In the photo, I can see a large black and white panda sitting on a tree branch. The image captures the panda's unique appearance and its natural habitat, showcasing the beauty and rarity of these animals.
- Analysis: The LMM's response is not false because it does not make any false claims about the image contents. For example, it does not claim that the panda is eating bamboo.
- Hallucination: No.

### Example 3:
- Image Contents: Fashion accessory, Mammal, Dog, Umbrella
- Question: Describe this image in details.
- Standard Human-Generated Answer: In this picture I can see a dog laying under a yellow umbrella on a beach, and there is a rope on the right side. In the background I can see few persons talking or lying. At the top of the image, I can see the blue and clear sky.
- LMM Response to Evaluate: The image features a brown dog laying on the sand at the beach, underneath a yellow umbrella. The dog is positioned near the water, and there are people in the background. The beach is sandy, and the sun is shining, creating a pleasant atmosphere. The dog is enjoying the shade provided by the umbrella, which is a common way to protect pets from the sun's harmful rays. The presence of people in the background suggests that the beach is a popular spot for leisure activities, such as sunbathing, swimming, and relaxing.
- Analysis: The LMM's response is false because it claims that the dog is positioned near the water, but there is no water in the image.
- Hallucination: Yes.

### Example 4:
- Image Contents: Clothing, Luggage and bags, Man, Building, Woman, Person
- Question: Where is this photo taken?
- Standard Human-Generated Answer: From the slogan "Welcome to Houston, Texas", we can infer that this photo is taken at an airport in Houston, Texas.
- LMM Response to Evaluate: This photo is taken at an airport, specifically at the entrance of a terminal. However, the specific location of this airport cannot be determined from the image.
- Analysis: Although the LMM's response does not explicitly mention the location "Houston, Texas" and does not perfectly answer the question, it makes no false claims about the image contents. Therefore, this response is not considered hallucination. It is OK if the LMM fails to directly answer the question, expresses uncertainty, or cannot include all the objects or details in the human-generated answer.
- Hallucination: No.

### Example 5:
- Image Contents: Tree, Tower
- Question: What environmental concerns are involved in this photo?
- Standard Human-Generated Answer: This photo shows a smokestack emitting smoke. It may cause severe air pollution which threatens human health.
- LMM Response to Evaluate: In the photo, there are several environmental concerns related to the smokestack emitting smoke. The smoke from the smokestack is a byproduct of industrial processes, which can contribute to air pollution, climate change, and negative impacts on human health. The smoke contains harmful pollutants, such as particulate matter, sulfur dioxide, nitrogen oxides, and carbon monoxide, which can lead to respiratory problems, heart disease, and other health issues. Additionally, the smoke contributes to the greenhouse effect, which can lead to global warming and climate change. The smokestack's emissions also affect the environment, as they can harm wildlife, vegetation, and ecosystems. Therefore, it is essential to address these environmental concerns by implementing measures to reduce emissions and promote sustainable practices in industrial processes.
- Analysis: Although the LMM's response is significantly longer than the standard human-generated answer, it does not contain any false claims about the image contents. Instead, it provides additional general information about the environmental concerns, which can be inferred from the smoke emission. Such detailed analysis or reasoning should be considered as a positive aspect, as long as it contains no false claims.
- Hallucination: No.

With these examples in mind, please help me evaluate whether the response by the LMM is informative, and whether hallucination exists in it, based on the comparison between the LMM's response and the factual information provided in the image contents, question, and the standard human-generated answer below.

Please note that the standard human-generated answer may only contain factual information but may not give a detailed analysis. Also, the standard human-generated answer may not be completely comprehensive in describing all the objects and their attributes, so please be a bit more cautious during evaluation. LMM's detailed analysis or reasoning should be encouraged.

To evaluate the LMM responses, first, begin your evaluation by providing a short explanation. Second, after providing your explanation, you must rate the response by choosing from the following options:
- Rating: 6, very informative with good analysis or reasoning, no hallucination
- Rating: 5, very informative, no hallucination
- Rating: 4, somewhat informative, no hallucination
- Rating: 3, not informative, no hallucination
- Rating: 2, very informative, with hallucination
- Rating: 1, somewhat informative, with hallucination
- Rating: 0, not informative, with hallucination

\end{lstlisting}